\documentclass[conference]{IEEEtran}
\IEEEoverridecommandlockouts
\usepackage{cite}
\usepackage{amsmath,amssymb,amsfonts}
\usepackage{algorithmic}
\usepackage{graphicx}
\graphicspath{ {images/} }
\usepackage{textcomp}
\def\BibTeX{{\rm B\kern-.05em{\sc i\kern-.025em b}\kern-.08em
    T\kern-.1667em\lower.7ex\hbox{E}\kern-.125emX}}
\begin{document}

\title{Automatic Coding for Neonatal Jaundice From Free Text Data Using Ensemble Methods
}

\author{\IEEEauthorblockN{Scott Werwath}
\IEEEauthorblockA{
\textit{UC Berkeley}\\
sbw@berkeley.edu}
}

\maketitle

\begin{abstract}
This study explores the creation of a machine learning model to automatically identify whether a Neonatal Intensive Care Unit (NICU) patient was diagnosed with neonatal jaundice during a particular hospitalization based on their associated clinical notes. We develop a number of techniques for text preprocessing and feature selection and compare the effectiveness of different classification models. We show that using ensembled decision tree classification, both with AdaBoost  and with bagging, outperforms support vector machines (SVM), the current state-of-the-art technique for neonatal jaundice coding.
\end{abstract}

\section{Introduction}
\subsection{Medical Coding}\label{AA}
In the course of normal clinical operations, healthcare providers are  required to \textit{code} patient records. The process of coding involves taking all data associated with a single patient case and abstracting out a set of standardized alphanumeric codes in which each unique code is associated with a unique diagnosis or medical procedure. These codes are used for a variety of applications, including medical billing, as labels in statistical analysis of medical datasets, as features in clinical decision making, and as tags in information retrieval systems. 

While many coding schemes exists, the most widely used is the International Classification of Diseases, 9\textsuperscript{th} Revision (ICD-9) \cite{Slee}. Each ICD-9 code, consisting of 3 to 5 alphanumeric characters, represents either a diagnosis or a procedure.

In actual clinical practice, coding is performed manually by professional medical coders. This process is both time-intensive and error-prone due to the fact that patient health records contain a large amount of data in free-text narrative format which is difficult to interpret quickly for both humans and computers \cite{Tange}\cite{OMalley}. 

A system for automatic coding would have the potential to substantially decrease the number of man-hours required to process clinical cases, but building such a system requires the ability to extract important information from free-text reports. To this end, the application of natural language processing (NLP) and machine learning (ML) techniques to clinical texts show much promise \cite{Meystre}.

\subsection{Neonatal Jaundice}\label{AA}
Neonatal jaundice, also called neonatal hyperbilirubinemia, is a yellow discoloration of the skin due to elevated bilirubin in a neonate's (infant's) blood. Transient neonatal jaundice is somewhat common in infants and usually carries little risk. However, untreated jaundice can put premature or otherwise ill infants at great risk of developing permanent neurological impairments \cite{Lantzy}. Additionally, it has been shown that the diagnosis of neonatal jaundice is a risk factor for both mortality and rehospitalization \cite{Escobar}. 

Both the risk factors associated with and the relative prevalence of neonatal jaundice make it an excellent subject for study using machine learning techniques; the risk factors provide the motivation, while the prevalence provides a wealth of data to be analyzed.

\section{Related Work}
The past few years have seen a sharp rise in the application of NLP in the medical field \cite{Wolniewicz}. The monetary and time cost of analyzing medical documents has motivated efforts to automate many manual tasks in the analysis and processing of medical texts.

Given the importance and time-intensive nature of medical coding, many scientists have started developing techniques for automated coding \cite{Meystre}. For most of these attempts, the task was narrow (i.e. attempting to predict a single ICD code or a small set of related codes) and utilized a relatively small dataset. 

At least one attempt has been made to use deep learning (namely character-level recurrent neural networks) to build more general coding models, but with limited success \cite{Shi}. The main limitation with this method is that deep learning models require a large amount of data to train \cite{Chen}, but even a large clinical dataset may only contain a few examples of case files with any given ICD code. Additionally, the ICD-9 standard contains over 14,000 different codes; treating coding as a simple multi-class classification problem with 14,000 different classes is infeasible.

More successful automated coding models have utilized non-deep machine learning techniques such as support vector machines (SVMs) \cite{Cortes} and focus on training a model to detect the presence of a single ICD code or class of related codes. Some of these methods also leverage structured data stored in health records in addition to free-text narratives \cite{Ferrao}.

At least one previous attempt has been made to build an automatic coding model for neonatal jaundice; \cite{Marafino} uses an SVM on the text of clinical notes. Our research improves on this model by employing an additional preprocessing step and using ensemble classification methods. We explore and compare these alternative classification techniques and analyze their efficacy.

No matter the exact method for extracting medical codes from free-text, two common NLP challenges must be solved. First, one must represent free-text in an appropriate feature space (usually as a vector). Second, one must train a model to accurately classify patient records based on the selected features. 

\section{Methodology}
\subsection{Dataset and Data Extraction}\label{AA}
The Medical Information Mart for Intensive Care III (MIMIC-III) dataset contains complete anonymized records from over 53,000 ICU stays \cite{Johnson}. These records were gathered from hospital records from the Beth Israel Deaconess Medical Center in Boston, Massachusetts between 2001 and 2012. Of these records, 7870 correspond to neonates admitted between 2001 and 2008. 

Each hospital stay in the MIMIC-III dataset is assigned a unique HADM\_ID identifier which can be used to cross-reference patient data back to their associated hospitalization.

In order to find all patients who were diagnosed with neonatal jaundice, we found all HADM\_IDs that were labeled with an ICD-9 code associated with neonatal jaundice. A list of such codes are shown in Table~\ref{tab1}. 
\begin{table}[htbp]
\caption{ICD-9 Codes for Neonatal Jaundice}
\begin{center}
\begin{tabular}{cc}
\textbf{ICD-9 Code}&\textbf{Description} \\
\hline
\textbf{773.0} & Hemolytic disease, RH isoimmunization \\
\textbf{773.1} & Hemolytic disease, ABO isoimmunization \\
\textbf{773.0} & Hemolytic disease, unknown \\
\textbf{774.1} & Perinatal jaundice, excessive hemolysis \\
\textbf{774.2} & Neonatal jaundice, preterm delivery \\
\textbf{774.30} & Neonatal jaundice, delayed conjugation \\
\textbf{774.31} &  Neonatal jaundice, delayed conjugation, other \\
\textbf{774.39} & Other jaundice \\
\textbf{774.6} & Unspecified fetal and neonatal jaundice \\
\end{tabular}
\label{tab1}
\end{center}
\end{table}
In total, 7092 hospitalizations were of neonates. Of these, 2868 stays were associated with a neonatal jaundice diagnosis and served as positive examples. We also selected the remaining 4224 non-jaundice NICU stays as negative examples.

For each relevant hospitalization, we concatenated all ICU notes, including nursing notes, radiology reports, physician progress notes, and discharge summaries into a single string of text called a \textit{noteset}. This string served as the raw data to be preprocessed and then classified. 
\subsection{Data Preprocessing}\label{AA}
Preprocessing of raw text is an extremely important element of any NLP model, especially in the task of classification \cite{Haddi}. In fact, the preprocessing of raw text is a form of manual feature selection, in which we are able to choose what parts of the document are to be given to the classifier \cite{Fatih}.

The first stage of our preprocessing algorithm is \textit{cleaning}, in which punctuation and other extraneous characters are stripped from the raw text. In addition to punctuation, numbers are also removed from the noteset. While numbers are often left in classifiers for medical texts, later cross-validation indicated that they were low-information for this particular task. Finally, we also remove anonymization indicators from the text of the document. These indicators are special flags left to indicate that confidential patient data had been redacted from the noteset at that point in the text, and generally contained information that was irrelevant for classification such as dates, names, and phone numbers.

After the noteset is cleaned, it is \textit{tokenized}, or split into a stream of individual words. The next series of preprocessing steps act on individual words or multiword phrases, and they work to reduce the \textit{vocabulary}, or number of unique words, of our dataset. Vocabulary reduction often results in improved classification performance \cite{Madsen} because the model does not have to learn to classify low-information words or learn that synonymous words or phrases should be classified similarly. Additionally, based on our representation of words as vectors (one-hot encodings), vocabulary reduction corresponds directly to a dimensionality reduction of samples.

The first stage of vocabulary reduction is \textit{semantic mapping}, i.e. the replacement of synonymous words or phrases with a standard token \cite{Banerjee}. This allows the model to learn only one representation of the underlying concept rather than having to learn many. For this, we use a list of 727 common medical words and phrases and use regular expressions to find and replace each of them in the notesets. Examples of such replacements can be found in Table~\ref{reptab}.
\begin{table}[htbp]
\caption{Example Replacements During Preprocessing}
\begin{center}
\begin{tabular}{ccc}
\textbf{Rule}&\textbf{Raw Text}&\textbf{Processed Text} \\
\hline
\textbf{Semantic Map} & "raises concern" & "RISK" \\
\textbf{Semantic Map} & "cannot rule out" & "RISK" \\
\textbf{Semantic Map} & "no evidence of" & "NEGEX" \\
\textbf{Semantic Map} & "rule out" & "NEGEX" \\
\textbf{Stemming} & "neonatal" & "neonat" \\
\textbf{Stemming} & "neonate" & "neonat" \\
\textbf{Stop Word Removal} & "the patient is" & "patient" \\
\textbf{Phrase Detection} & "heart attack" & "heart\_attack" \\
\end{tabular}
\label{reptab}
\end{center}
\end{table}

After semantic mapping, we apply \textit{stemming}, which is the process of reducing words to their root morpheme (see Table~\ref{reptab}). For this task, we use the Snowball stemming algorithm \cite{Porter}. Since different morphological variations of the same word (e.g. "neonatal" and "neonate") likely have similar meanings, stemming allows the model to learn  a single classification for all morphological forms for a word.

Next, \textit{stop words} were removed from the notesets. Stop words are common words such as "the" and "it" which convey no real semantic information. The removal of stop words from raw text has been shown to improve the performance of classification models due to their being low-information and useless for most classification tasks \cite{Silva}. 

Then, collocations, or common multiword phrases, are detected and replaced with a single token. In order to automatically detect collocations in our corpus, we calculate the \textit{normalized pointwise mutual information} (NPMI) of every pair of words in the corpus \cite{Bouma}. The NPMI is given by \eqref{pmi}, where $x$ and $y$ are indicators for a two given words appearing in text, $p(x,y)$ is the probability of the two words appearing next to each other, and all probabilities are measured empirically from our corpus.
\begin{equation}
i_n(x,y) = \frac{-1}{\ln p(x,y)}\ln \frac{p(x,y)}{p(x)p(y)}\label{pmi}
\end{equation}
Intuitively, NMPI measures how much the observed co-occurance of two words $x$ and $y$ differs from what we would expect if their occurrences were independent of each other. Token pairs in our corpus with high NMPI were normalized into a single token (see Table~\ref{reptab}).

Finally, the text was searched for phrases such as "no" and "rule out" that indicate negation. We assume that the three words following such phrases are negated, and so we remove them from the document. Since our bag of words representation fails to take word order into account, this prevents the model from interpreting the word "jaundice" in the phrase "rule out jaundice" as indicating a positive result. 
\subsection{Document Representation}\label{AA}
In order to classify notesets, we have to represent them as vectors. While deeper methods (e.g. convolutional neural networks) are able to take word order into account, we choose to use a \textit{bag-of-words} (BoW) model in which we ignore the actual ordering of words in a document and instead simply count the number of occurrences of each unique word in the document. This has the benefit of greatly simplifying our document representation, but it throws away potentially valuable information contained in word ordering (e.g. "negative for cancer, positive for flu" and "positive for cancer, negative for flu" will have the same BoW representation). Despite this disadvantage, BoW representations work very well in practice for many applications.

A naive approach to BoW representation is a \textit{count vector}. In a count vector, each unique word is represented by a $V$-dimensional one hot encoding vector, where $V$ is the size of the vocabulary. A document is then represented by the sum (or sometimes average) of the one hot encoding of all the words in the document. The result is that, at each index in a document vector, the entry at index $i$ is the number of times the $i$\textsuperscript{th} word in the vocabulary occurs in the document.

A commonly used technique to improve upon the count vector representation to weight each entry by the term's \textit{inverse document frequency} (IDF), given in \eqref{idf}. $N$ is the total number of documents in the corpus, and the denominator represents the number of documents in the corpus where the word $t$ appears.
\begin{equation}
idf(t) = \log \frac{N}{|\{d \in D : t \in d\} |}\label{idf}
\end{equation}
The IDF measures the rarity of a word in a given corpus. From an information theoretic perspective, words with high IDF (i.e. rarer words) provide more information, so we want to give them more weight in our decision making process.

When we weight each count vector entry by its word's IDF and normalize appropriately, we get that each entry $t$ in the vector representation for document $d$ is given by \eqref{tfidf}.
\begin{equation}
tfidf(t, d) = \frac{|t' \in d : t' = t|}{|d|} idf(t)\label{tfidf}
\end{equation}
This value is called the \textit{term frequency-inverse document frequency} (TF-IDF), and a vector of these values is called a TF-IDF vector.

\subsection{Baseline SVM Model}\label{AA}
Reference \cite{Marafino} demonstrates the use of a support vector machine (SVM) in automatic coding for neonatal jaundice. This study used a grid search to determine that a linear kernel with penalty hyperparameter of $C=100$ was ideal. This model was reimplemented, and our own grid search over relevant hyperparmeters with 10-fold cross-validation found those same settings to be optimal. This SVM model was used as a baseline to compare with the performance of new ensemble learning methods. 
\subsection{Ensemble Methods}\label{AA}
\textit{Ensemble learning} refers to the practice of using multiple machine learning models and combining their individual outputs. Ensemble methods often provide better results than using a single model because they can reduce variance. When a set of models are trained with random samples of a dataset or on a random subset of features, we are essentially sampling from the space of possible classifiers. By taking many samples and averaging their results, we reduce our reliance on the particularities of any one classifier and therefore reduce variance \cite{Dietterich}.

We used the sklearn package \cite{Pedregosa} to implement two ensembled classifiers: AdaBoost with decision trees \cite{Zhu} and bagged decision trees \cite{Breiman}. 

The base estimator of both ensemble methods are \textit{decision trees}. A decision tree uses a tree-like model in which each node represents a decision based on the value of a single feature, and each subsequent branch represents a path chosen based on that decision. Trees are constructed such that each decision boundary maximally decreases the Gini Impurity of the two sets of samples which would fall on either side of the boundary. Put another way, each decision node of the trees attempts to best separate the training data based on the value of a single feature. For our implementation of decision trees, the depth of the tree is allowed to grow indefinitely until all leaves contain pure subsets of the training set. 

Decision trees tend to suffer from problems such as instability and overfitting \cite{Gareth}. One technique to counter such problems is \textit{bagging}, in which $n$ different decision trees (in our case, $n = 10$ was found to be optimal) are trained on different samples of the training data. To classify a new sample, each of the $n$ trees performs classification, and then a vote is taken to determine the final label.

\textit{AdaBoost} is another ensemble method which attempts to improve on the performance of a single base classifier. Rather than sampling uniformly from the training set to train each classifier, each decision tree is trained on the same training set, but with different weights for each training point for each classifier. The values of the weights for classifier $k$ are determined by how poorly the first $k-1$ classifiers predict the point, such that greater weight is given to misclassified points. To classify a new sample, each individual trees performs classification, and then a weighted vote is taken to determine the final label. Using cross-validation, we found that using 40 boosted decision trees worked best for our problem.

\section{Results}
A summary of the performance of SVM and the two ensemble classifiers is presented in Table~\ref{restab}.
\begin{table}[htbp]
\caption{Performance of Various Classifiers}
\begin{center}
\begin{tabular}{ccccc}
\textbf{Classifier}&\textbf{Accuracy}&\textbf{Precision}&\textbf{Recall}&\textbf{F\textsubscript{1} Score} \\
\hline
\textbf{SVM} & 0.865 & 0.835 & 0.839 & 0.837 \\
\textbf{Boosted Trees} & 0.907 & \textbf{0.923} & 0.846 & 0.883 \\
\textbf{Bagged Trees} & \textbf{0.909} & 0.904 & \textbf{0.876} & \textbf{0.890} \\
\end{tabular}
\label{restab}
\end{center}
\end{table}
The F\textsubscript{1} score refers to the harmonic average of precision and recall, given in \eqref{f1}.
\begin{equation}
F_1 = \frac{2}{precision^{-1} + recall^{-1}}\label{f1}
\end{equation}

Our reimplementation of the SVM model in \cite{Marafino} did not perform as well as the original, though this is likely due to differences in the dataset. Both of our ensemble methods outperform the two SVM implementations in every performance metric. 

\begin{figure}[htbp]
\centerline{\includegraphics[scale=0.6]{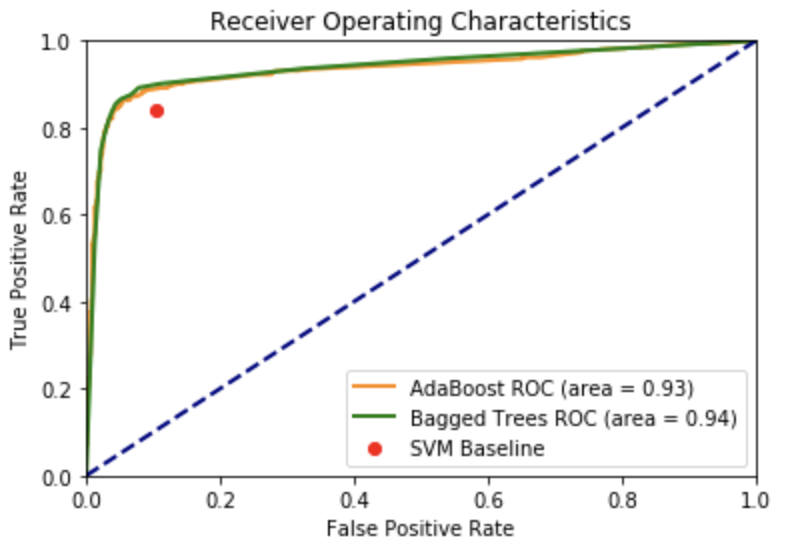}}
\caption{ROC curves of ensemble methods compared to SVM baseline}
\label{roc}
\end{figure}

The receiver operating characteristic (ROC) curves of the ensemble methods are shown in in Fig.~\ref{roc}. We see that the baseline SVM performance lies under the the ROC curves of both ensemble methods, demonstrating that the ensemble methods are able to outperform SVM. 

Finally, an AdaBoost classifier was trained on only discharge summaries, rather than entire notesets. The accuracy of the classifier decreased from 0.91 to 0.86, implying that non-discharge notes contained information relevant to the classification.
\section{Discussion}
\subsection{Performance of Ensemble Methods}
Both the AdaBoost and bagged trees classifiers outperformed the baseline SVM. While we were unable to find statistics on the accuracy of human medical coders in coding for neonatal jaundice, the average accuracy of humans in coding for a large variety of diagnoses seems to range roughly from 0.90 to 0.95 \cite{OMalley}. Therefore, it is possible that our models are approaching human-level performance at the task of coding for neonatal jaundice.

\begin{figure}[htbp]
\centerline{\includegraphics[scale=0.6]{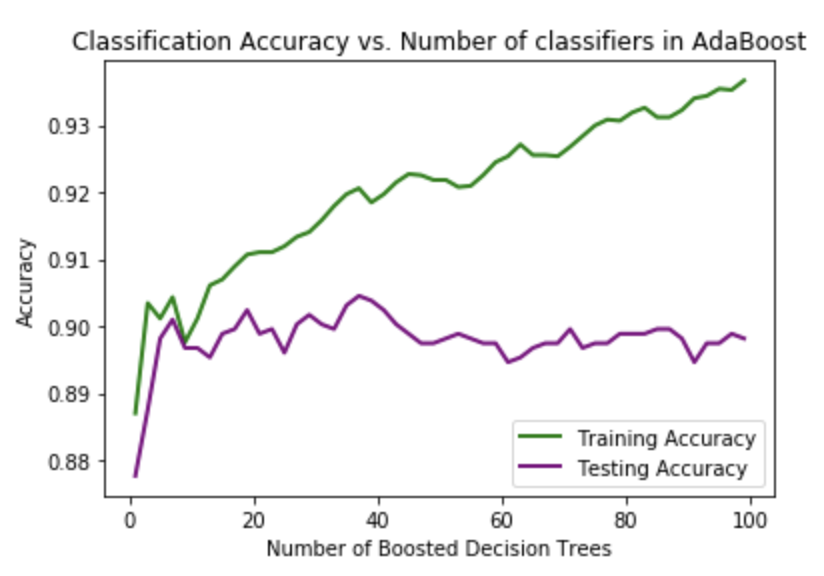}}
\caption{The effect of adding more classifiers on AdaBoost accuracy. Overfitting occurs when more than 40 classifiers are used.}
\label{num}
\end{figure}

When training ensemble methods, one important hyperparameter is the number of estimators to train. Fig.~\ref{num} shows the performance of AdaBoost as a function of the number of boosted decision tree estimators.
We see that adding more than 40 classifiers increases training accuracy while having no effect on testing accuracy, demonstrating that overfitting is occurring.

In the task of medical detection, recall is often considered a more important metric than precision, since false positives can be manually checked and corrected by a human while false negatives may go unnoticed. As such, the bagged trees model may be favored for actual use in a clinical setting.
\subsection{Feature Importance}
\begin{figure*}[htbp]
\centerline{\includegraphics[scale=0.5]{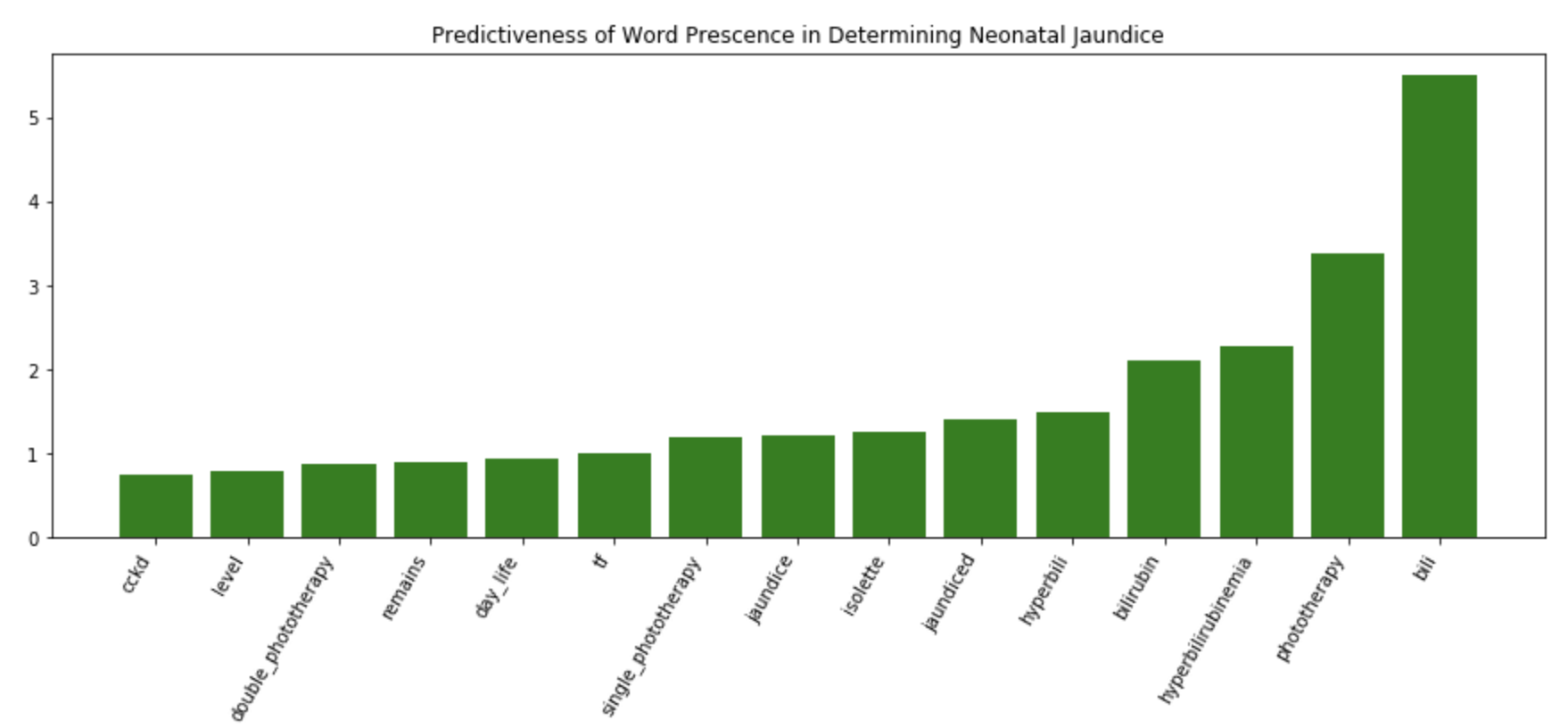}}
\caption{Relevance of words in determining a positive SVM classification}
\label{rel}
\end{figure*}
While SVM was outperformed by ensemble methods, it can still provide interesting insight into what words are most likely to indicate neonatal jaundice.

In SVM with a linear kernel, the learned classifier is the ideal separating hyperplane. This hyperplane, like all hyperplanes, can be uniquely represented by a single vector which is orthogonal to the hyperplane; we will call this the coefficient vector. Since features that are orthogonal (or close to orthogonal) to the hyperplane are most relevant to classification, we can interpret the entries of the coefficient vector as a metric of feature importance, where large positive entries indicate that a feature contributes greatly to a positive classification \cite{Guyon}.

In TF-IDF vectors, each dimension (i.e. feature) corresponds to a single word's occurrence in that noteset. Therefore, that feature's corresponding coefficient measures the predictiveness of that word's presence to a diagnosis of neonatal jaundice. We plot these most-relevant terms in Fig.~\ref{rel}.
We see that the most relevant features include terms such as "bili", an abbreviation for bilirubin (the chemical which causes jaundice), and "phototherapy", the standard treatment for patients with jaundice. Based on this manual inspection of the top features, it seems that SVM is learning to classify notesets based on words that are relevant to neonatal jaundice. 
\section{Conclusion}
We have shown that both bagged decision trees and boosted decision trees are able to outperform SVM at the task of automatically detecting a neonatal jaundice diagnosis from clinical notes and reports; in fact, the ensemble methods approach may be approaching human-level performance. The ability of these models to achieve high performance based entirely on free text data implies that the clinical notes contain sufficient information to carry out the task of assigning ICD codes. As more labeled clinical text becomes available, training highly accurate coding models will become feasible for a wider set of ICD codes.

\end{document}